\documentclass[11pt]{article}

\usepackage[preprint]{acl}
\usepackage{times}
\usepackage{latexsym}
\usepackage{booktabs}
\usepackage{multirow}
\usepackage[T1]{fontenc}
\usepackage[utf8]{inputenc}
\usepackage{float}
\usepackage{amsmath}

\usepackage{xcolor}
\usepackage{subcaption}
\usepackage{colortbl}
\definecolor{parallelColor}{HTML}{6D6E71} 
\definecolor{oursColor}{HTML}{C41230}     
\colorlet{Standardtint}{parallelColor!12!white}
\colorlet{s3tint}{oursColor!10!white}
\usepackage{microtype}
\usepackage{inconsolata}
\usepackage{graphicx}
\usepackage{tcolorbox}
\tcbuselibrary{skins, breakable}
\usepackage{paracol}
\usepackage{amssymb}      
\usepackage{pifont}       

\definecolor{neutralBox}{HTML}{F5F5F5}     
\definecolor{poolBox}{HTML}{FAFAFA}        
\colorlet{winBg}{s3tint}                 

\newcommand{\cmark}{\textcolor{green!50!black}{\ding{51}}}
\newcommand{\xmark}{\textcolor{red!70!black}{\ding{55}}}

\newtcolorbox{casebox}{
  colback=white, colframe=black!40,
  boxrule=0.5pt, arc=2pt,
  left=8pt, right=8pt, top=6pt, bottom=6pt,
  before skip=8pt, after skip=4pt
}

\newtcolorbox{diagbox}{
  colback=neutralBox, colframe=black!20,
  boxrule=0.4pt, arc=2pt,
  left=8pt, right=8pt, top=5pt, bottom=5pt,
  before skip=6pt, after skip=6pt
}

\newtcolorbox{poolbox}{
  colback=poolBox, colframe=black!15,
  boxrule=0.3pt, arc=2pt,
  left=8pt, right=8pt, top=4pt, bottom=4pt,
  before skip=6pt, after skip=6pt,
  fontupper=\footnotesize
}

\definecolor{threadBox}{HTML}{FCFCFC}      
\definecolor{threadFrame}{HTML}{D8D8D8}    

\newtcolorbox{thr}[1]{%
  enhanced,
  colback=threadBox,
  colframe=threadFrame,
  boxrule=0.4pt,
  arc=2pt,
  left=6pt, right=6pt, top=4pt, bottom=4pt,
  before skip=4pt, after skip=4pt,
  fonttitle=\bfseries\footnotesize,
  title=#1,
  coltitle=black,
  attach boxed title to top left={xshift=6pt, yshift=-2pt},
  boxed title style={
    colback=white,
    colframe=threadFrame,
    boxrule=0.4pt,
    arc=2pt,
    left=4pt, right=4pt, top=1pt, bottom=1pt,
  },
  fontupper=\footnotesize,
}

\newcommand{\sq}[1]{%
  \par\noindent\textcolor{black!60}{$\hookrightarrow$ search:}~\texttt{\small #1}\par
}
\newcommand{\sr}[1]{%
  \par\noindent\textcolor{black!60}{$\hookleftarrow$ result:}~\textit{\small #1}\par
}
\newcommand{\note}[1]{%
  \par\noindent\textcolor{black!50}{\scriptsize #1}\par
}
\newcommand{\ans}[1]{%
  \par\noindent\textbf{Answer:}~\texttt{#1}\par
}
\newcommand{\answin}[1]{%
  \par\noindent\colorbox{winBg}{\textbf{Answer:}~\texttt{#1}}\par
}

\newcommand{\colhdr}[3]{%
  \noindent\colorbox{#3}{%
    \parbox{0.97\linewidth}{%
      \centering\textbf{#1}\quad\small(QPD = #2)%
    }%
  }\par\medskip
}

\newcommand{\oraclecmark}[1]{\par\noindent\textbf{Oracle:}~\cmark~\textit{\small #1}\par}
\newcommand{\oraclexmark}[1]{\par\noindent\textbf{Oracle:}~\xmark~\textit{\small #1}\par}

\definecolor{midnightgreen}{rgb}{0.0, 0.29, 0.33}

\title{Beyond Parallel Sampling: Diverse Query Initialization for Agentic Search}

\author{ \bf
Sidhaarth Murali$^{a}$, 
João Coelho$^{a,b}$, 
Jingjie Ning$^{a}$, \\ \bf
João Magalhães$^c$,   
Bruno Martins$^b$,  
Chenyan Xiong$^a$\\
$^a$ Carnegie Mellon University, United States \\
$^b$ Instituto Superior Técnico and INESC-ID, University of Lisbon, Portugal\\
$^c$ NOVA LINCS, NOVA School of Science and Technology, Portugal \\
\normalsize{ssmurali@andrew.cmu.edu}
}

\begin{document}
\maketitle
\begin{abstract}
Test-time scaling for agentic search typically increases depth (i.e., more turns and tokens per
trajectory) or breadth (i.e., more parallel rollouts). Here we focus on breadth scaling, showing
that standard parallel sampling yields diminishing returns, tracing this to query redundancy
at the first turn. When models issue similar first queries across rollouts, the threads retrieve
overlapping evidence, and subsequent turns are conditioned on this shared retrieval. We address
this limitation with \textbf{\texttt{DivInit}}, a training-free intervention at the first turn. Rather than
sampling $k$ independent first queries, \textbf{\texttt{DivInit}} draws $n$ candidates from a
single call, picks $k < n$ diverse seeds, and runs them as parallel trajectories. Across five
open-weight models and eight benchmarks, \textbf{\texttt{DivInit}} consistently improves over
standard parallel sampling, with average gains of five to seven points on multi-hop QA at matched
compute.
\footnote{\raggedright
Code available at \href{https://github.com/cxcscmu/diverse-query-initialization}{diverse-query-initialization}.
}
\end{abstract}

\section{Introduction}
\label{sec:intro}

Test-time scaling has become one of the most reliable ways to improve language model performance on complex tasks without training. On search agents, two practical levers dominate current work. The first, depth, extends a single trajectory through longer chains of thought, more reasoning turns, or more search-and-read iterations~  \citep{wei2022cot,muennighoff2025s1, yao2022react,li2025searcho1}. The second, breadth, runs multiple trajectories in parallel and aggregates their outputs through voting, best-of-$N$ selection, or related strategies~\citep{wang2023selfconsistency, cobbe2021training,
brown2024large, snell2024scaling}.

\begin{figure}[t]
    \centering
    \hspace*{-0.04\columnwidth}%
    \includegraphics[width=1\columnwidth]{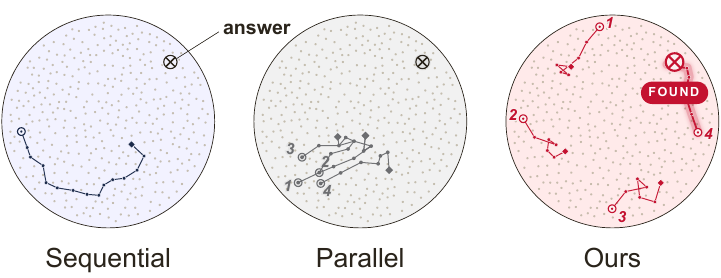}
    \caption{Conceptual illustration of retrieval-path exploration under sequential execution, standard parallel sampling, and our approach named \textbf{\texttt{DivInit}}.}
    \label{fig:evidence_space}
    \vspace{-0.15cm}
\end{figure}

In this work, we focus on parallel test-time scaling of search agents (i.e., breadth). We empirically identify an \textit{anchor collapse} phenomenon in multi-turn agentic search, where the turn-1 query anchors the trajectory and the $k$ parallel threads collapse onto a single retrieval path. When threads start with similar first queries, overlapping evidence is retrieved, and systems fail in correlated ways. Parallel sampling still outperforms a single sequential trajectory~\citep{li2026benchmark}, but the low diversity hints at a suboptimal usage of resources (Figure~\ref{fig:evidence_space}).

While standard parallel scaling launches $k$ independent threads that often anchor on the same turn-1 query, one can instead choose a diverse starting set before any thread runs. The rest of the search agent would be unchanged, making this intervention compatible with any agentic search loop. We name this approach \textbf{\texttt{DivInit}}, sampling $n$ candidate queries from a single shared call, picking $k < n$  through Maximal Marginal Relevance (MMR)~\citep{carbonell1998mmr}, and running one trajectory per chosen query.

We evaluate \textbf{\texttt{DivInit}} on five open-weight models across eight benchmarks covering multi-hop QA and long-horizon reasoning. Results show that \textbf{\texttt{DivInit}} consistently improves over standard parallel sampling at matched compute, with average gains of five to seven points on multi-hop QA. The improvement grows with model size on most datasets, suggesting a capacity floor below which models cannot act productively on varied first queries. Moreover, diversity at turn 1 is sufficient on its own, and extending the pool selection to later turns yields little benefit, since the first turn separation carries through subsequent retrieval.

\section{Related Work}
\label{sec:related}

\paragraph{Test-time Scaling.}
Scaling inference-time compute without retraining has proven effective
across reasoning and retrieval tasks, through approaches such as
chain-of-thought prompting \citep{wei2022cot}, budget forcing
\citep{muennighoff2025s1}, process-reward supervision
\citep{lightman2023verify,uesato2022solving}, self-consistency
\citep{wang2023selfconsistency}, or best-of-$N$ selection
\citep{cobbe2021training}. For long-horizon agentic tasks, AggAgent
\citep{lee2026agentic} shows that treating parallel trajectories as an
interactive environment outperforms voting and answer-concatenation
baselines. \textbf{\texttt{DivInit}} addresses a different part of the
same problem: rather than improving aggregation after trajectories
complete, it diversifies where they start.

\paragraph{Agentic Search.}
ReAct \citep{yao2022react} introduced the thought-action loop for
tool-augmented agents, and IRCoT \citep{trivedi2023ircot} extended this to
multi-hop retrieval. A separate line trains agents via outcome or
process rewards \citep{jin2025searchr1,chen2025research,
song2025r1searcher,sun2025zerosearch, wen2026smartsearch, anonymous2026recycle}. \textbf{\texttt{DivInit}} changes only the distribution from which the first query
is drawn, before the loop runs, and is compatible with any of the above.


\paragraph{Diversity in Generation and Retrieval.}
Diverse beam search \citep{vijayakumar2018diverse}, nucleus sampling \citep{holtzman2020nucleus}, maximal marginal relevance \citep{carbonell1998mmr}, and determinantal point processes \citep{kulesza2012dpp,chen2018dpp} promote diversity within a single generation or retrieval call. \textbf{\texttt{DivInit}} operates across parallel agent threads at the query level, directly spreading the retrieval neighborhoods each thread will explore.
\section{Anchor Collapse in Agentic Search}
\label{sec:anchor}

\begin{figure}[t]
    \centering
    \begin{subfigure}[b]{0.49\columnwidth}
        \includegraphics[width=\textwidth]{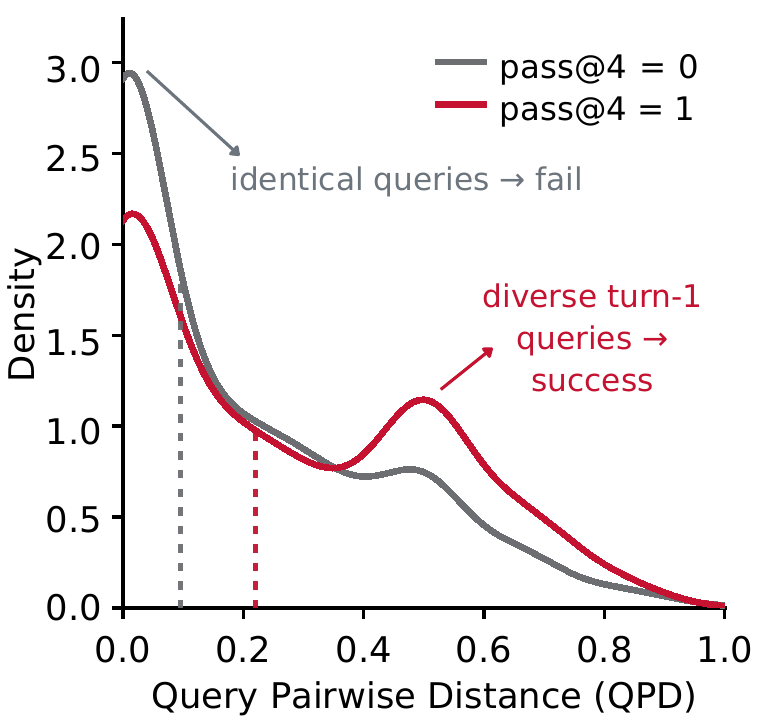}
        \label{fig:qpd_passk}
    \end{subfigure}
    \hfill
    \begin{subfigure}[b]{0.49\columnwidth}
        \includegraphics[width=\textwidth]{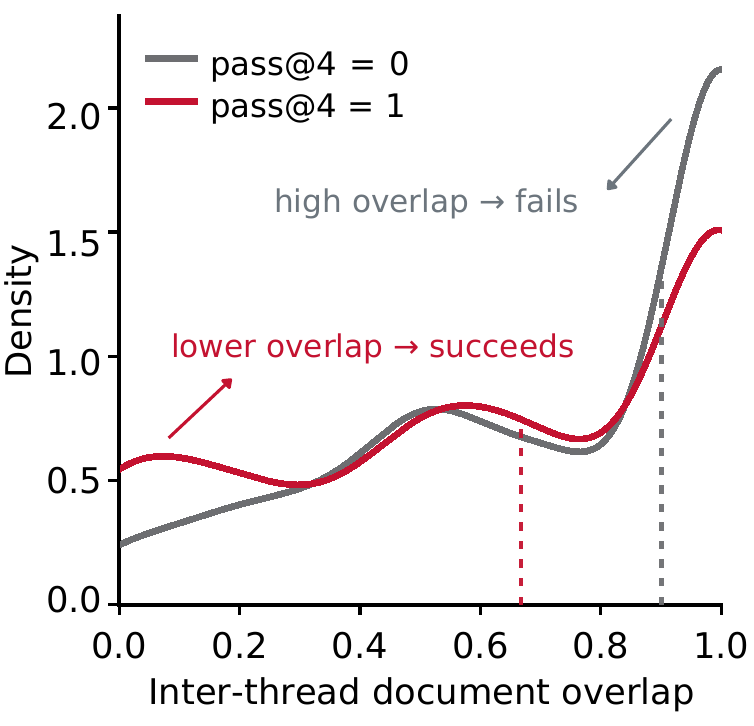}
        \label{fig:overlap_passk}
    \end{subfigure}
    \caption{
        Turn-1 QPD and inter-thread document overlap under standard
        parallel sampling (Qwen3-8B, $k{=}4$, $\tau{=}1.0$), split
        by pass@4 outcome. Failed questions concentrate at low QPD
        and high overlap; successful questions show higher QPD and
        lower overlap.
    }
    \label{fig:passk_diagnostics}
\end{figure}

\paragraph{Agent Setup.}
We build on the ReAct-style search agent of \citet{li2026benchmark}.  
Given a question $q$, the agent iteratively issues search queries, retrieves the top-$N$ documents, and decides whether to continue searching or commit to an answer. Each trajectory runs for at most $T$ turns. Standard parallel sampling draws $k$ independent trajectories for the same question under temperature $\tau$.

\paragraph{Anchor Collapse.}
To characterize interaction diversity across the $k$ trajectories of a question, we measure two quantities. \textbf{Query Pairwise Distance (QPD)} is the mean pairwise Jaccard distance among tokens from turn-1 queries, where higher QPD indicates that trajectories initialize from distinct retrieval directions. \textbf{Document overlap} is the mean pairwise Jaccard similarity between retrieved document sets, where higher overlap indicates that trajectories consume largely identical evidence despite surface-level query variation.

A preliminary study of standard parallel sampling on the GAIA benchmark~\citep{DBLP:conf/iclr/MialonF0LS24} shows that these quantities track pass@4 outcome (Figure~\ref{fig:passk_diagnostics}). Questions where all trajectories fail (pass@4\,${=}$\,0) cluster at low QPD and high document overlap, whereas questions with at least one successful trajectory (pass@4\,${=}$\,1) show a secondary mode with higher QPD and lower overlap.

We hypothesize that this behavior arises from the sequential dependency structure of agentic retrieval. Because each turn conditions on previously retrieved evidence, trajectories with similar turn-1 retrieval actions tend to remain coupled throughout the search process. Early retrieval decisions therefore anchor subsequent reasoning and evidence acquisition, causing the $k$ trajectories to collapse onto a narrow region of the search space. We refer to this phenomenon as \textbf{anchor collapse}.

As an example, consider a GAIA question asking for the NASA award number supporting a specific author's work, linked from a Universe Today article. Under standard parallel sampling, all four trajectories issue identical queries packing every entity into one search
(turn-1 QPD 0.098), retrieve the same wrong award number from a different paper, and exhaust their budget without finding the target
record. A full trace is provided in Appendix~\ref{sec:appendix-cases}.

\section{\textbf{\texttt{DivInit}}: Diverse Query Initialization}
\label{sec:method}

The diagnostic in Section \ref{sec:anchor} identified low turn-1 QPD as a predictor of failure, i.e., threads that issue near-identical first queries tend to retrieve overlapping evidence and collapse onto a single trajectory. Next we describe \textbf{\texttt{DivInit}}, which aims to control this behavior by oversampling queries in the first turn, and selecting a diverse subset.


\paragraph{Procedure.}
The intervention is restricted to the first turn. While standard parallel sampling has each of $k$ threads generate its own turn-1 query independently, we draw a pool $\mathcal{C}$ of $n > k$ candidate queries in a single LLM call at temperature $\tau$. We select $k$ from $\mathcal{C}$ via MMR \citep{carbonell1998mmr}, initializing the selected set $S$ with the pair of candidates at maximum pairwise distance and adding subsequent picks as follows:
\begin{equation}
\label{eq:mmr}
\small
c^{\star} = \operatorname*{arg\,max}_{c \in \mathcal{C} \setminus S} \left[ (1 - \lambda) \min_{s \in S} d_J(c, s) - \lambda\, d_J(c, q) \right],
\end{equation}
where $d_J$ is the token-level Jaccard distance, and $q$ is the original question. We stop when $|S| = k$ and run one thread per selected candidate.

\paragraph{Compute Overhead.}
The shared pool replaces the $k$ independent turn-1 LLM calls of standard sampling with a single call that emits $n$ candidates in one decoding pass, consolidating $k$ prefills into one at the cost of a longer output. From turn 2, the two methods are identical, so the total cost is $1 + k(T{-}1)$ calls, $k{-}1$ fewer than standard's $kT$.

\section{Experimental Setup}
\label{sec:setup}

We evaluate on five open-weight models: Qwen3 (1.7B, 4B, 8B)
\citep{yang2025qwen3} and Gemma3 (4B, 12B)
\citep{gemmateam2025gemma3technicalreport}, all served locally via vLLM~\citep{kwon2023efficient}.
Benchmarks span two groups. First, multi-hop question answering (HotpotQA,
MuSiQue, 2WikiMHQA, Bamboogle, FRAMES), for which the agent searches over a local Wiki18 BM25 index~\citep{jin2025FlashRAG}. Second, larger models are also evaluated on open-web reasoning (GAIA, HLE, WebWalker), searching the live web via SERPER~\citep{serper2025api}. We sample 500 questions per benchmark, except for GAIA which uses the full 103-question validation split.

The agent runs $T{=}8$ turns at $k{=}4$ threads and temperature $\tau{=}1.0$. Retrieval returns the textual content of the top 10 documents per query. For \textbf{\texttt{DivInit}} (Eq.~\ref{eq:mmr}), we draw $n{=}16$ candidates in the first turn, and set $\lambda = 0$ for MMR. Prompts are detailed in Appendix~\ref{app:prompts}.

\begin{table*}[t]
\centering
\tiny
\setlength{\tabcolsep}{2pt}
\renewcommand{\arraystretch}{1.15}
\begin{tabular}{l cc cc cc cc cc cc !{\vrule width 0.5pt} cc cc cc cc}
\toprule
& \multicolumn{12}{c}{\textbf{Multi-hop QA (Wiki18)}}
& \multicolumn{8}{c}{\textbf{Open-Web Reasoning (Serper)}} \\
\cmidrule(lr){2-13}\cmidrule(lr){14-21}
& \multicolumn{2}{c}{\textbf{HpQA}}
& \multicolumn{2}{c}{\textbf{MuSi}}
& \multicolumn{2}{c}{\textbf{2Wiki}}
& \multicolumn{2}{c}{\textbf{Bambo}}
& \multicolumn{2}{c}{\textbf{FRAMES}}
& \multicolumn{2}{c}{\textbf{Avg}}
& \multicolumn{2}{c}{\textbf{GAIA}}
& \multicolumn{2}{c}{\textbf{HLE}}
& \multicolumn{2}{c}{\textbf{WebWalker}}
& \multicolumn{2}{c}{\textbf{Avg}} \\
\cmidrule(lr){2-3}\cmidrule(lr){4-5}\cmidrule(lr){6-7}\cmidrule(lr){8-9}\cmidrule(lr){10-11}\cmidrule(lr){12-13}\cmidrule(lr){14-15}\cmidrule(lr){16-17}\cmidrule(lr){18-19}\cmidrule(lr){20-21}
\textbf{Model}
& \cellcolor{Standardtint}\textbf{S} & \cellcolor{s3tint}\textbf{DI}
& \cellcolor{Standardtint}\textbf{S} & \cellcolor{s3tint}\textbf{DI}
& \cellcolor{Standardtint}\textbf{S} & \cellcolor{s3tint}\textbf{DI}
& \cellcolor{Standardtint}\textbf{S} & \cellcolor{s3tint}\textbf{DI}
& \cellcolor{Standardtint}\textbf{S} & \cellcolor{s3tint}\textbf{DI}
& \cellcolor{Standardtint}\textbf{S} & \cellcolor{s3tint}\textbf{DI}
& \cellcolor{Standardtint}\textbf{S} & \cellcolor{s3tint}\textbf{DI}
& \cellcolor{Standardtint}\textbf{S} & \cellcolor{s3tint}\textbf{DI}
& \cellcolor{Standardtint}\textbf{S} & \cellcolor{s3tint}\textbf{DI}
& \cellcolor{Standardtint}\textbf{S} & \cellcolor{s3tint}\textbf{DI} \\
\midrule
Qwen3-1.7B
& \cellcolor{Standardtint}42.9 & \cellcolor{s3tint}\textbf{43.8}
& \cellcolor{Standardtint}14.5 & \cellcolor{s3tint}\textbf{15.6}
& \cellcolor{Standardtint}37.6 & \cellcolor{s3tint}\textbf{41.5}
& \cellcolor{Standardtint}16.8 & \cellcolor{s3tint}\textbf{24.3}
& \cellcolor{Standardtint}13.1 & \cellcolor{s3tint}\textbf{13.6}
& \cellcolor{Standardtint}$25.0 \pm1.3 $ & \cellcolor{s3tint}$\mathbf{27.8} \pm1.1 $
& \cellcolor{Standardtint}- & \cellcolor{s3tint}-
& \cellcolor{Standardtint}- & \cellcolor{s3tint}-
& \cellcolor{Standardtint}- & \cellcolor{s3tint}-
& \cellcolor{Standardtint}- & \cellcolor{s3tint}- \\
Qwen3-4B
& \cellcolor{Standardtint}41.9 & \cellcolor{s3tint}\textbf{53.2}
& \cellcolor{Standardtint}15.9 & \cellcolor{s3tint}\textbf{19.7}
& \cellcolor{Standardtint}41.9 & \cellcolor{s3tint}\textbf{49.0}
& \cellcolor{Standardtint}32.5 & \cellcolor{s3tint}\textbf{40.8}
& \cellcolor{Standardtint}15.5 & \cellcolor{s3tint}\textbf{20.4}
& \cellcolor{Standardtint}$29.5 \pm1.0 $ & \cellcolor{s3tint}$\mathbf{36.6} \pm1.3 $
& \cellcolor{Standardtint}22.7 & \cellcolor{s3tint}\textbf{27.8}
& \cellcolor{Standardtint}9.7 & \cellcolor{s3tint}\textbf{14.3}
& \cellcolor{Standardtint}38.7 & \cellcolor{s3tint}\textbf{44.9}
& \cellcolor{Standardtint}$23.7 \pm1.6 $ & \cellcolor{s3tint}$\mathbf{29.0} \pm2.7 $ \\
Qwen3-8B
& \cellcolor{Standardtint}50.4 & \cellcolor{s3tint}\textbf{57.0}
& \cellcolor{Standardtint}23.9 & \cellcolor{s3tint}\textbf{29.7}
& \cellcolor{Standardtint}46.3 & \cellcolor{s3tint}\textbf{55.1}
& \cellcolor{Standardtint}47.7 & \cellcolor{s3tint}\textbf{57.6}
& \cellcolor{Standardtint}24.8 & \cellcolor{s3tint}\textbf{30.8}
& \cellcolor{Standardtint}$38.6 \pm1.2 $ & \cellcolor{s3tint}$\mathbf{46.0} \pm1.3 $
& \cellcolor{Standardtint}26.0 & \cellcolor{s3tint}\textbf{30.2}
& \cellcolor{Standardtint}10.0 & \cellcolor{s3tint}\textbf{14.1}
& \cellcolor{Standardtint}41.6 & \cellcolor{s3tint}\textbf{46.8}
& \cellcolor{Standardtint}$25.2 \pm0.8 $ & \cellcolor{s3tint}$\mathbf{28.2} \pm2.6 $ \\
\midrule
Gemma3-4B
& \cellcolor{Standardtint}40.0 & \cellcolor{s3tint}\textbf{49.2}
& \cellcolor{Standardtint}\textbf{17.2} & \cellcolor{s3tint}16.1
& \cellcolor{Standardtint}42.8 & \cellcolor{s3tint}\textbf{52.2}
& \cellcolor{Standardtint}27.7 & \cellcolor{s3tint}\textbf{37.9}
& \cellcolor{Standardtint}12.3 & \cellcolor{s3tint}\textbf{14.7}
& \cellcolor{Standardtint}$28.0 \pm1.0 $ & \cellcolor{s3tint}$\mathbf{34.0} \pm 1.1 $
& \cellcolor{Standardtint}- & \cellcolor{s3tint}-
& \cellcolor{Standardtint}- & \cellcolor{s3tint}-
& \cellcolor{Standardtint}- & \cellcolor{s3tint}-
& \cellcolor{Standardtint}- & \cellcolor{s3tint}- \\
Gemma3-12B
& \cellcolor{Standardtint}54.9 & \cellcolor{s3tint}\textbf{59.1}
& \cellcolor{Standardtint}31.6 & \cellcolor{s3tint}\textbf{36.1}
& \cellcolor{Standardtint}52.0 & \cellcolor{s3tint}\textbf{53.9}
& \cellcolor{Standardtint}55.7 & \cellcolor{s3tint}\textbf{64.3}
& \cellcolor{Standardtint}31.0 & \cellcolor{s3tint}\textbf{37.5}
& \cellcolor{Standardtint}$45.0 \pm1.1 $ & \cellcolor{s3tint}$\mathbf{50.2} \pm0.9$
& \cellcolor{Standardtint}34.0 & \cellcolor{s3tint}\textbf{35.0}
& \cellcolor{Standardtint}12.7 & \cellcolor{s3tint}\textbf{14.8}
& \cellcolor{Standardtint}38.0 & \cellcolor{s3tint}\textbf{45.2}
& \cellcolor{Standardtint}$28.2 \pm2.2 $ & \cellcolor{s3tint}$\mathbf{31.6 \pm 1.4}$ \\
\bottomrule
\end{tabular}
\caption{
Pass@4 (\%) results obtained with a mean over 3 seeds (with standard deviation reported on group averages only). Per-dataset columns show standard parallel sampling (S) vs \textbf{\texttt{DivInit}} (DI). Left: multi-hop QA over a local Wiki18 index. Right: open-web reasoning via SERPER.
}
\label{tab:results}
\end{table*}

\section{Experiments}
\label{sec:results}

In this section, we measure the empirical gains of \textbf{\texttt{DivInit}} over standard parallel sampling, and analyze the query-level patterns that drive them.

\begin{figure}[t]
    \centering
    \includegraphics[width=\columnwidth]{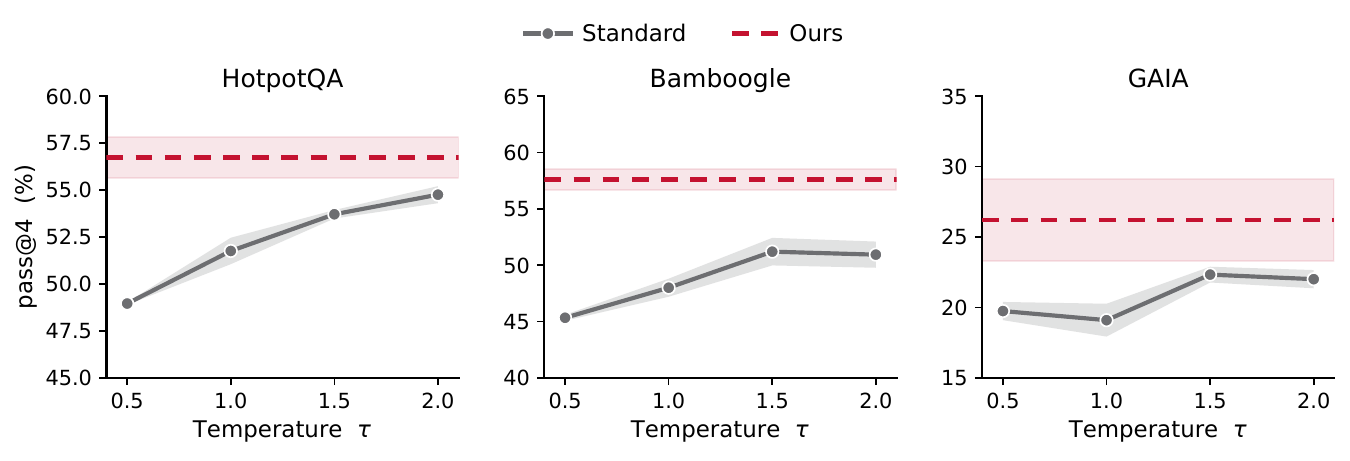}
\caption{Pass@4 under standard parallel sampling at varying $\tau$ (Qwen3-8B, $k{=}4$), with \textbf{\texttt{DivInit}} at $\tau{=}1.0$ as reference. Higher temperature improves the baseline, but does not close the gap.}
    \label{fig:temperature_ablation}
\end{figure}

\subsection{Empirical Results}

Table~\ref{tab:results} compares \textbf{\texttt{DivInit}} against standard parallel sampling under a fixed token budget. Results show that \textbf{\texttt{DivInit}} improves on nearly every cell. On the open-web group, WebWalker carries the largest improvements (six to seven points across all three models). While the table shows pass@$4$, the results also hold for $k=8$ (Figures~\ref{fig:appfig-mhqa-passk} and~\ref{fig:appfig-reasoning-passk} of Appendix~\ref{app:metrics-grid}) and carry through to pass@1 aggregation under AggAgent~\citep{lee2026agentic} (Appendix~\ref{app:aggregation}). \textbf{\texttt{DivInit}} only differs from the standard parallel scaling strategy on the first turn. A wall-clock time comparison of both is provided in Appendix~\ref{app:wall_time}. 

Results also show that improvement scales with model size. Qwen3 averages move from $\Delta{=}2.8$ at 1.7B to $\Delta{=}7.4$ at 8B. Notably, the 1.7B numbers point to a capacity floor below which query diversification yields limited benefits.

Figure~\ref{fig:temperature_ablation} sweeps $\tau \in \{0.5, 1.0, 1.5, 2.0\}$ for standard parallel sampling, with \textbf{\texttt{DivInit}} at $\tau{=}1.0$ as reference. Performance tends to improve with temperature, but saturates below \textbf{\texttt{DivInit}}, confirming that sampling noise is not a substitute for explicit diversity selection.

\subsection{Qualitative Analysis}
\label{sec:analysis}

\begin{figure}[t]
    \centering
    \includegraphics[width=0.49\columnwidth]{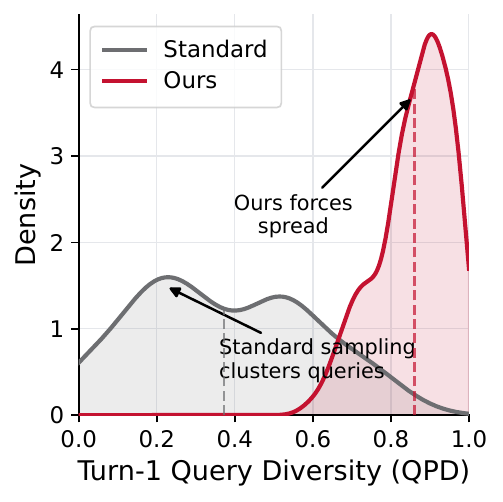}
    \hfill
    \includegraphics[width=0.49\columnwidth]{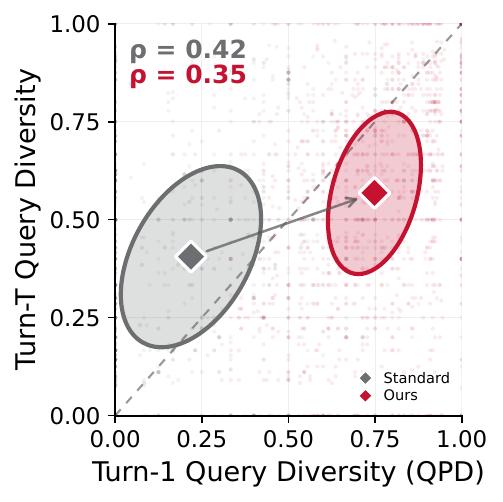}
\caption{Turn-1 QPD distribution (left) and turn-1 QPD vs.\ ATD per question (right), Qwen3-8B, $k{=}4$. \textbf{\texttt{DivInit}} shifts queries toward high diversity, and this separation persists across all turns.}
    \label{fig:qualitative}
\vspace{-0.15cm}
\end{figure}

Figure~\ref{fig:qualitative} shows that standard parallel sampling concentrates turn-1 queries near low QPD values ($\approx 0.2$), while \textbf{\texttt{DivInit}} shifts them toward high diversity ($\approx 0.85$). To test if this separation persists beyond the first turn, we measure Across-Thread Distance (ATD), i.e. the mean pairwise Jaccard distance between tokens in queries from different threads, pooled across all turns.

Questions with higher turn-1 QPD tend to exhibit higher ATD, with positive correlation in both conditions ($\rho_{\text{std}}{=}0.42$, $\rho_{\text{DivInit}}{=}0.35$). This further solidifies that turn-1 retrieval acts as a trajectory anchor, i.e., once threads retrieve different evidence early, subsequent reasoning and search remain separated with little additional intervention. Figures~\ref{fig:appfig-mhqa-diag} and~\ref{fig:appfig-reasoning-diag} of Appendix~\ref{app:metrics-grid} show this pattern across the full grid of benchmarks and models. 
\subsection{Ablations}
\label{sec:ablations}

\paragraph{Extent of diversification.}
\textbf{\texttt{DivInit}} applies pool selection only at turn 1. With Qwen3-8B, extending it to turns 1 through $N$ for $N \in \{1,...,8\}$ yields no gain on the open-web benchmarks (Table~\ref{tab:extent}). This matches the observations from \S\ref{sec:analysis}, where early trajectory separation persists throughout later retrieval turns without additional intervention.

\paragraph{Diversity strategy.}
\textbf{\texttt{DivInit}} selects $k$ queries from the pool of $n$ via MMR with $\lambda$=0, which maximizes the minimum pairwise distance in the selected set. On GAIA with Qwen3-8B, this scores 34.0 pass@4. Setting $\lambda \in \{0.5, 0.75\}$ adds weight to similarity between each candidate and the original question, and drops performance to 30--31, indicating that the LLM-generated pool is already on-topic. Uniform random selection from
the same pool scores 27.2, below all MMR variants.

\begin{table}[t]
\centering
\tiny
\setlength{\tabcolsep}{3pt}
\renewcommand{\arraystretch}{1.1}
\begin{tabular}{ll cccccccc}
\toprule
& & \multicolumn{8}{c}{\textbf{Turn} $N$} \\
\cmidrule(lr){3-10}
\textbf{Model} & \textbf{Dataset} & 1 & 2 & 3 & 4 & 5 & 6 & 7 & 8 \\
\midrule
Qwen3-8B
& GAIA      & \textbf{30.7} & 25.9 & 27.8 & 30.7 & 28.2 & 28.2 & 28.2 & 27.8 \\
& HLE       & \textbf{13.9} & 12.1 & 11.6 & 11.1 & 12.4 & 13.1 & 13.5 & 13.3 \\
& WebWalker & \textbf{47.1} & 45.3 & \textbf{47.1} & 46.3 & 45.1 & 44.3 & 44.8 & 44.8 \\
\bottomrule
\end{tabular}
\caption{Pass@4 (\%) results as \textbf{\texttt{DivInit}} pool selection extends to turns 1 through $N$.}
\label{tab:extent}
\vspace{-0.3cm}
\end{table}

\section{Conclusions and Future Work}
\label{sec:conclusion}

We identified anchor collapse as a failure mode of standard parallel sampling within agentic
search systems, where $k$ threads converge onto near-identical first queries, retrieve overlapping
evidence, and fail in correlated ways. We also proposed \textbf{\texttt{DivInit}} as a
training-free fix that oversamples a shared candidate pool of queries and selects a maximally
diverse subset before any thread runs. Across five open-weight models and eight benchmarks, the
\textbf{\texttt{DivInit}} intervention improves over standard parallel sampling at matched compute,
with gains of five to seven points on multi-hop QA.

The results suggest that standard breadth scaling underutilizes inference-time compute by
allocating rollout budget to correlated trajectories. The same issue may arise in RL training:
in group-based methods such as GRPO, near-identical search trajectories produce low-variance
rollout groups and weak learning signals~\citep{singh2026rl, anonymous2026recycle}. Diversifying trajectory
initialization before group formation is a natural next step. A complementary direction is
aggregation. \textbf{\texttt{DivInit}} maximizes thread pool diversity but reports pass@$k$,
while exploiting the structural diversity of the pool to produce a single answer remains open.

\section*{Limitations}
\label{sec:limitations}

The pass@$k$ metric corresponds to the ceiling of what the thread pool can achieve, not the accuracy of any single answer the system would produce in deployment. We also report single-answer accuracy under AggAgent~\citep{lee2026agentic} in Appendix~\ref{app:aggregation} and find that diversification gains transfer to pass@1, but closing the remaining gap to pass@$k$ is an open problem.

Anchor collapse is characterized here for search agents, where the turn-1 decision is a query string and its downstream effect is observable through retrieved documents. Whether the same failure mode arises in other domains, such as tool-use and code, is not addressed in this work.

\section*{Ethical Considerations}

All the evaluation datasets and pre-trained models used in our experiments are publicly available for research usage. We will provide the source code that allows for reproduction of the results.

By using large pre-trained language models, we acknowledge inherent biases embedded within the models, which may perpetuate or amplify societal biases present in the training data.
\bibliography{custom}

\clearpage
\appendix
\clearpage
\section{Reproducibility Details}
\label{sec:appendix-reproducibility}

\subsection{Prompts}
\label{app:prompts}

We use two agent prompts and one pool-generation prompt. All three are
reproduced verbatim below (placeholders in curly braces are filled at
run time). The multi-hop QA agent uses a ReAct loop with
\texttt{<thought>}, \texttt{<search>}, and \texttt{<answer>} actions.
The open-web agent adds a \texttt{<summary>} action to compress prior
\texttt{<information>} blocks within longer trajectories. The
pool-generation prompt is invoked once at turn 1 in every
\textbf{\texttt{DivInit}} run to produce the candidate pool of
$n{=}16$ queries.

\begin{tcolorbox}[breakable,colback=gray!5,colframe=gray!40,fonttitle=\bfseries,
  title=Multi-hop QA agent prompt,boxrule=0.5pt]
\small\ttfamily
You are a research assistant that answers questions by searching the
web. You have \{max\_turns\} turns total. You are on turn \{turn\}.\\[2pt]
Question: \{question\}\\[2pt]
History of searches and findings:\\
\{history\}\\[2pt]
Rules (follow strictly):\\
1. Every response must use exactly one of these two formats.\\
2. If you need more information: output \texttt{<thought>...</thought>}
then \texttt{<search>your single search query here</search>}.\\
3. If you have the answer (or it is the last turn): output
\texttt{<thought>...</thought>} then
\texttt{<answer>your concise final answer here</answer>}.\\
4. Do not output an answer outside \texttt{<answer>...</answer>}.
Do not output search and answer in the same turn.\\
5. On the last turn (turn \{max\_turns\}) you MUST output
\texttt{<answer>...</answer>} with your best guess if unsure.
\end{tcolorbox}

\begin{tcolorbox}[breakable,colback=gray!5,colframe=gray!40,fonttitle=\bfseries,
  title=Open-Web agent prompt,boxrule=0.5pt]
\small\ttfamily
You are a precise research assistant answering web-reasoning questions
by searching the internet. You have \{max\_turns\} turns total. You
are on turn \{turn\}.\\[2pt]
Question: \{question\}\\[2pt]
Search history:\\
\{history\}\\[2pt]
Available actions per turn (choose exactly one):\\
\hspace*{1em}\texttt{<search>query</search>}~~-- issue a new search\\
\hspace*{1em}\texttt{<summary>...</summary>}~~-- compress prior
\texttt{<information>} blocks\\
\hspace*{1em}\texttt{<answer>...</answer>}~~-- submit your final
answer\\[2pt]
Rules:\\
1. Use EXACTLY one action per response.\\
2. Answers must be exact: a name, number, date, or short phrase.\\
3. On turn \{max\_turns\} you MUST give \texttt{<answer>...</answer>}.\\
4. Search for specific facts. Avoid generic queries.
\end{tcolorbox}

\begin{tcolorbox}[breakable,colback=gray!5,colframe=gray!40,fonttitle=\bfseries,
  title=Pool-generation prompt,boxrule=0.5pt]
\small\ttfamily
Generate exactly \{n\} diverse search queries to investigate this
question. Each query should approach the question from a different
angle, specifically targeting different constraints or components of
the question.\\
\{history\_block\}\\[2pt]
Question: \{question\}\\[2pt]
Output exactly \{n\} queries, one per line, numbered 1--\{n\}.
No other text.
\end{tcolorbox}

\subsection{Evaluation Protocol}
\label{app:judge}

All numbers reported in the main paper are evaluated by \texttt{gpt-4o-mini}
(temperature 0, max 300 output tokens). For each question, \text{pass@}$k$ is 1 if at least
one of the $k$ candidates is correct, and 0 otherwise. We report the mean over questions.
Because the same judge is applied to every condition, any bias affects all methods equally.
The total judge cost is approximately \$175.

\begin{tcolorbox}[breakable,colback=gray!5,colframe=gray!40,fonttitle=\bfseries,
  title=Judge prompt (multi-hop QA),boxrule=0.5pt]
\small\ttfamily
You are an expert evaluator. Determine if the generated answer
correctly answers the question based on the ground truth answer.\\[2pt]
Question: \{question\}\\
Ground Truth Answer: \{ground\_truth\}\\
Generated Answer: \{generated\_answer\}\\[2pt]
Evaluation Rubric:\\
1. Factuality: the answer must contain the core correct information.\\
2. Semantic equivalence: mark CORRECT if the meaning is the same
despite phrasing differences.\\
3. Completeness: for multi-part questions, all parts must be correctly
answered.\\
4. Contradiction: mark INCORRECT only if the answer directly
contradicts the ground truth.\\
5. Extra information: ignore extra details if the core answer is
correct.\\[2pt]
Briefly explain your reasoning, then output ``CORRECT'' or
``INCORRECT'' on the final line.
\end{tcolorbox}

\begin{table*}[t]
\centering
\small
\setlength{\tabcolsep}{3pt}
\renewcommand{\arraystretch}{1.15}
\begin{tabular}{l | cc cc | cc cc}
\toprule
& \multicolumn{2}{c}{\textbf{Pass@4}}
& \multicolumn{2}{c}{\textbf{AggAgent@1}}
& \multicolumn{2}{c}{\textbf{Pass@4}}
& \multicolumn{2}{c}{\textbf{AggAgent@1}} \\
\cmidrule(lr){2-3}\cmidrule(lr){4-5}\cmidrule(lr){6-7}\cmidrule(lr){8-9}
\textbf{Dataset}
& \cellcolor{Standardtint}\textbf{S} & \cellcolor{s3tint}\textbf{DI}
& \cellcolor{Standardtint}\textbf{S} & \cellcolor{s3tint}\textbf{DI}
& \cellcolor{Standardtint}\textbf{S} & \cellcolor{s3tint}\textbf{DI}
& \cellcolor{Standardtint}\textbf{S} & \cellcolor{s3tint}\textbf{DI} \\
\midrule
\multicolumn{5}{l}{\textit{Qwen3-4B}} & \multicolumn{4}{l}{\textit{Qwen3-8B}} \\
HotpotQA
& \cellcolor{Standardtint}$41.9_{\pm0.6}$ & \cellcolor{s3tint}$\mathbf{53.2}_{\pm0.5}$
& \cellcolor{Standardtint}$38.5_{\pm0.2}$ & \cellcolor{s3tint}$\mathbf{46.7}_{\pm0.7}$
& \cellcolor{Standardtint}$49.3_{\pm2.8}$ & \cellcolor{s3tint}$\mathbf{56.2}_{\pm2.0}$
& \cellcolor{Standardtint}$44.6_{\pm2.3}$ & \cellcolor{s3tint}$\mathbf{47.8}_{\pm2.1}$ \\
MuSiQue
& \cellcolor{Standardtint}$15.9_{\pm0.8}$ & \cellcolor{s3tint}$\mathbf{19.7}_{\pm1.0}$
& \cellcolor{Standardtint}$13.0_{\pm0.8}$ & \cellcolor{s3tint}$\mathbf{13.6}_{\pm1.5}$
& \cellcolor{Standardtint}$24.2_{\pm0.9}$ & \cellcolor{s3tint}$\mathbf{31.3}_{\pm2.6}$
& \cellcolor{Standardtint}$13.6_{\pm0.4}$ & \cellcolor{s3tint}$\mathbf{17.4}_{\pm1.3}$ \\
2WikiMultihopQA
& \cellcolor{Standardtint}$41.9_{\pm0.3}$ & \cellcolor{s3tint}$\mathbf{49.0}_{\pm1.0}$
& \cellcolor{Standardtint}$29.0_{\pm0.3}$ & \cellcolor{s3tint}$\mathbf{35.5}_{\pm1.3}$
& \cellcolor{Standardtint}$46.8_{\pm1.3}$ & \cellcolor{s3tint}$\mathbf{55.0}_{\pm1.2}$
& \cellcolor{Standardtint}$35.2_{\pm1.7}$ & \cellcolor{s3tint}$\mathbf{38.9}_{\pm0.9}$ \\
Bamboogle
& \cellcolor{Standardtint}$32.5_{\pm1.5}$ & \cellcolor{s3tint}$\mathbf{40.8}_{\pm1.7}$
& \cellcolor{Standardtint}$30.9_{\pm3.4}$ & \cellcolor{s3tint}$\mathbf{37.1}_{\pm1.2}$
& \cellcolor{Standardtint}$47.3_{\pm2.3}$ & \cellcolor{s3tint}$\mathbf{57.8}_{\pm1.3}$
& \cellcolor{Standardtint}$40.4_{\pm0.9}$ & \cellcolor{s3tint}$\mathbf{46.1}_{\pm1.3}$ \\
FRAMES
& \cellcolor{Standardtint}$15.5_{\pm0.3}$ & \cellcolor{s3tint}$\mathbf{20.3}_{\pm1.3}$
& \cellcolor{Standardtint}$14.5_{\pm0.3}$ & \cellcolor{s3tint}$\mathbf{19.1}_{\pm1.1}$
& \cellcolor{Standardtint}$24.9_{\pm0.4}$ & \cellcolor{s3tint}$\mathbf{30.5}_{\pm1.1}$
& \cellcolor{Standardtint}$20.4_{\pm0.9}$ & \cellcolor{s3tint}$\mathbf{22.5}_{\pm0.2}$ \\
\midrule
\multicolumn{5}{l}{\textit{Gemma3-4B}} & \multicolumn{4}{l}{\textit{Gemma3-12B}} \\
HotpotQA
& \cellcolor{Standardtint}$40.0_{\pm0.8}$ & \cellcolor{s3tint}$\mathbf{49.2}_{\pm0.3}$
& \cellcolor{Standardtint}$33.4_{\pm0.2}$ & \cellcolor{s3tint}$\mathbf{39.2}_{\pm0.3}$
& \cellcolor{Standardtint}$54.9_{\pm1.0}$ & \cellcolor{s3tint}$\mathbf{59.0}_{\pm1.2}$
& \cellcolor{Standardtint}$46.2_{\pm0.3}$ & \cellcolor{s3tint}$\mathbf{50.4}_{\pm1.0}$ \\
MuSiQue
& \cellcolor{Standardtint}$\mathbf{17.2}_{\pm1.3}$ & \cellcolor{s3tint}$16.1_{\pm1.1}$
& \cellcolor{Standardtint}$\mathbf{17.0}_{\pm0.6}$ & \cellcolor{s3tint}$15.3_{\pm0.6}$
& \cellcolor{Standardtint}$31.6_{\pm0.9}$ & \cellcolor{s3tint}$\mathbf{36.1}_{\pm0.2}$
& \cellcolor{Standardtint}$22.8_{\pm0.3}$ & \cellcolor{s3tint}$\mathbf{30.2}_{\pm0.3}$ \\
2WikiMultihopQA
& \cellcolor{Standardtint}$42.8_{\pm0.7}$ & \cellcolor{s3tint}$\mathbf{52.1}_{\pm0.6}$
& \cellcolor{Standardtint}$18.6_{\pm2.2}$ & \cellcolor{s3tint}$\mathbf{34.3}_{\pm2.4}$
& \cellcolor{Standardtint}$52.0_{\pm1.0}$ & \cellcolor{s3tint}$\mathbf{53.9}_{\pm1.1}$
& \cellcolor{Standardtint}$\mathbf{36.1}_{\pm0.2}$ & \cellcolor{s3tint}$\mathbf{36.1}_{\pm1.0}$ \\
Bamboogle
& \cellcolor{Standardtint}$27.7_{\pm1.0}$ & \cellcolor{s3tint}$\mathbf{37.9}_{\pm1.9}$
& \cellcolor{Standardtint}$20.8_{\pm0.2}$ & \cellcolor{s3tint}$\mathbf{31.6}_{\pm2.6}$
& \cellcolor{Standardtint}$55.7_{\pm1.0}$ & \cellcolor{s3tint}$\mathbf{64.3}_{\pm0.4}$
& \cellcolor{Standardtint}$46.1_{\pm1.4}$ & \cellcolor{s3tint}$\mathbf{55.7}_{\pm3.1}$ \\
FRAMES
& \cellcolor{Standardtint}$12.3_{\pm0.1}$ & \cellcolor{s3tint}$\mathbf{14.6}_{\pm0.5}$
& \cellcolor{Standardtint}$12.1_{\pm0.7}$ & \cellcolor{s3tint}$\mathbf{13.2}_{\pm1.0}$
& \cellcolor{Standardtint}$30.9_{\pm0.2}$ & \cellcolor{s3tint}$\mathbf{37.5}_{\pm0.6}$
& \cellcolor{Standardtint}$25.8_{\pm0.6}$ & \cellcolor{s3tint}$\mathbf{28.9}_{\pm0.7}$ \\
\bottomrule
\end{tabular}
\caption{
    Pass@4 and AggAgent pass@1 aggregation, comparing standard parallel (S) against
    \textbf{\texttt{DivInit}} (DI) at $k{=}4$.
}
\label{tab:agg}
\vspace{-0.25cm}
\end{table*}

\subsection{Infrastructure and Reproducibility}
Open-weight models are served locally with vLLM in bf16 with prefix
caching enabled on $4{\times}$ NVIDIA L40S GPUs (48~GB VRAM). Per-rollout
seeds are derived deterministically from a SHA-256 hash of the question
identifier, the run-level seed, and the rollout index. Search backends
are queried with up to five retries on transient failures, and retrieved
text is truncated at $4{,}000$ characters per turn. Total compute is
approximately 590 GPU-hours plus \$185 in API costs. The Wiki18 backend
is fully reproducible, although SERPER search results drift over time, producing 0.1
to 0.3 points of drift on pass@$k$ across re-runs.

\label{app:hyperparams}

\section{Additional Empirical Results}
\label{sec:appendix-data}

\subsection{Pool Size}
\label{app:pool}

Table~\ref{tab:poolsize} shows results when varying the oversampled pool size $n$,
reporting pass@4 values. The performance is essentially unchanged across $n \in \{8,16,32\}$, while $n{=}4$ shows a drop, suggesting the pool should be larger than $k$.
\begin{table}[t]
\centering
\setlength{\tabcolsep}{6pt}
\renewcommand{\arraystretch}{1.1}
\small
\begin{tabular}{ll cccc}
\toprule
\textbf{Model} & \textbf{Dataset} & $n{=}4$ & $n{=}8$ & $n{=}16$ & $n{=}32$ \\
\midrule
\multirow{2}{*}{Qwen3-1.7B}
& HpQA & 42.6 & 44.1 & 44.3 & 44.4 \\
& GAIA & 6.5  & 7.1  & 7.1  & 7.4  \\
\addlinespace[1pt]
\multirow{2}{*}{Qwen3-8B}
& HpQA & 50.6 & 55.8 & 54.5 & 54.0 \\
& GAIA & 26.5 & 30.0 & 30.6 & 30.3 \\
\bottomrule
\end{tabular}
\caption{Pass@4 (\%) under \textbf{\texttt{DivInit}} as pool size varies.}
\vspace{-0.25cm}
\label{tab:poolsize}
\end{table}

\subsection{Performance Curves}
\label{app:metrics-grid}

\begin{figure*}[p]
\centering
\includegraphics[width=0.75\textwidth]{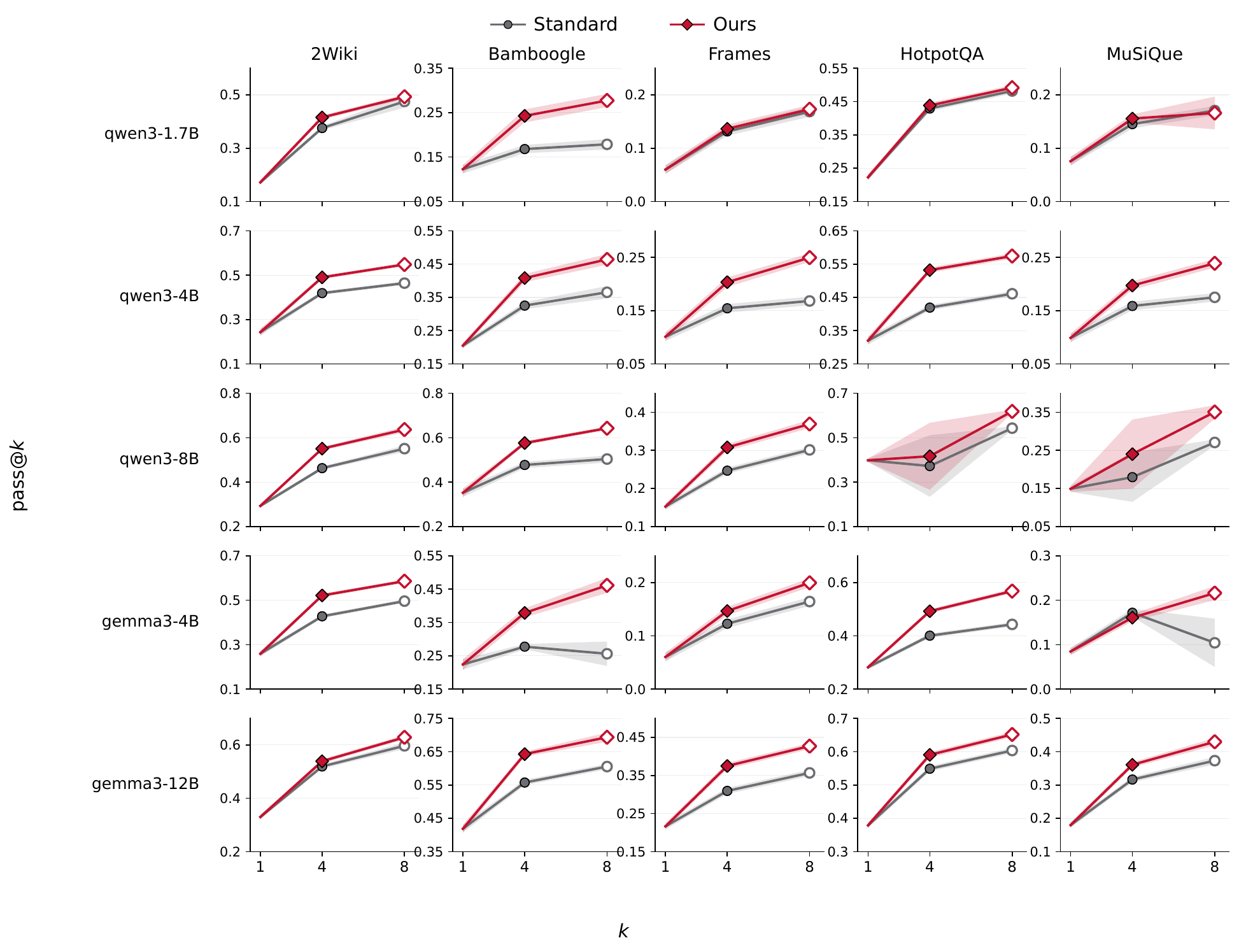}
\caption{Multi-hop QA results, reporting pass@$k$ as $k$ varies from 1 to 8, across five models and five datasets. \textbf{\texttt{DivInit}} retains its advantage at every $k$ under matched compute.}
\label{fig:appfig-mhqa-passk}
\end{figure*}

\begin{figure*}[p]
\centering
\includegraphics[width=0.75\textwidth]{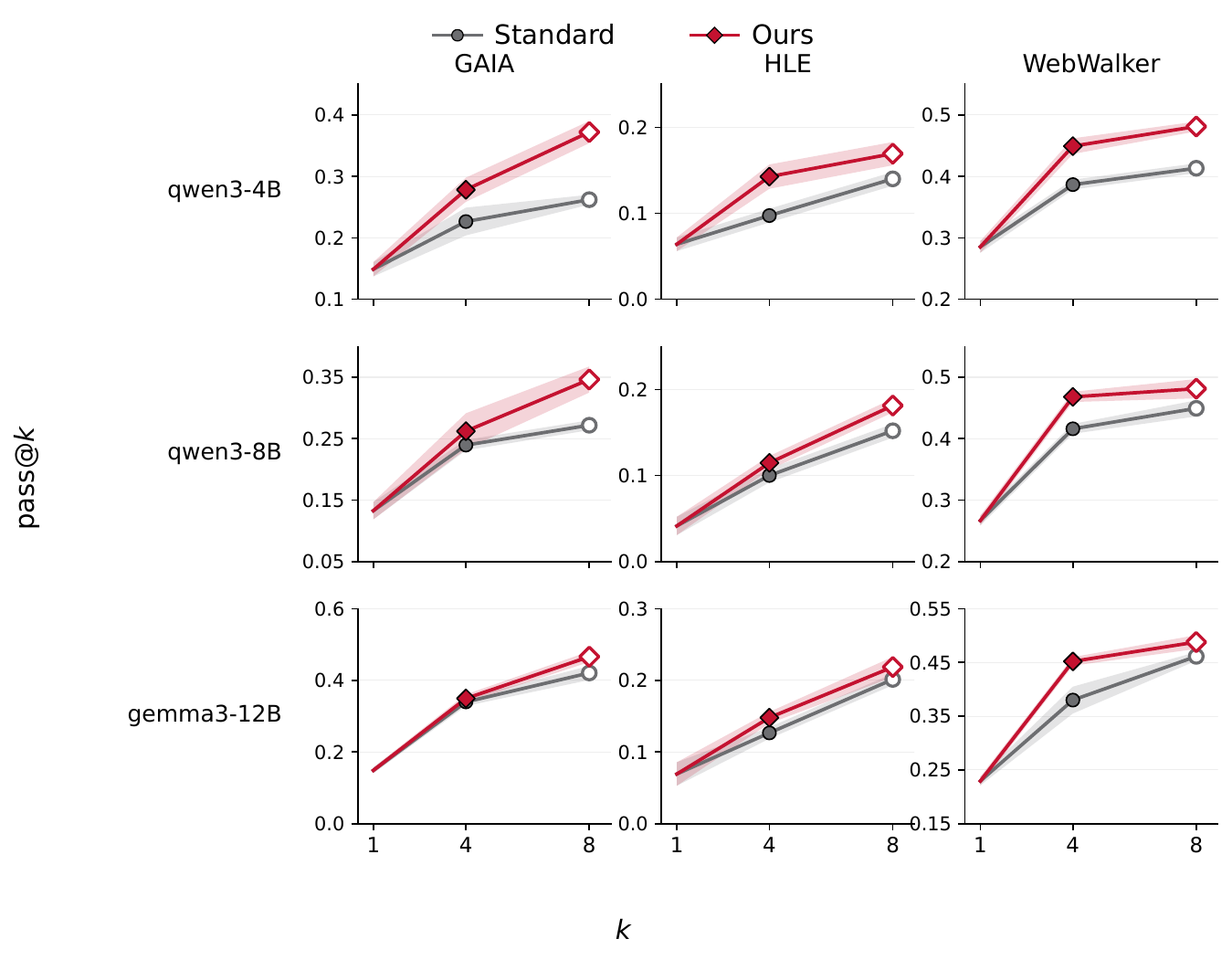}
\caption{Open-web reasoning results, reporting pass@$k$ as $k$ varies from 1 to 8. Layout matches
Figure~\ref{fig:appfig-mhqa-passk}.}
\label{fig:appfig-reasoning-passk}
\end{figure*}

\begin{figure*}[p]
\centering
\includegraphics[width=0.75\textwidth]{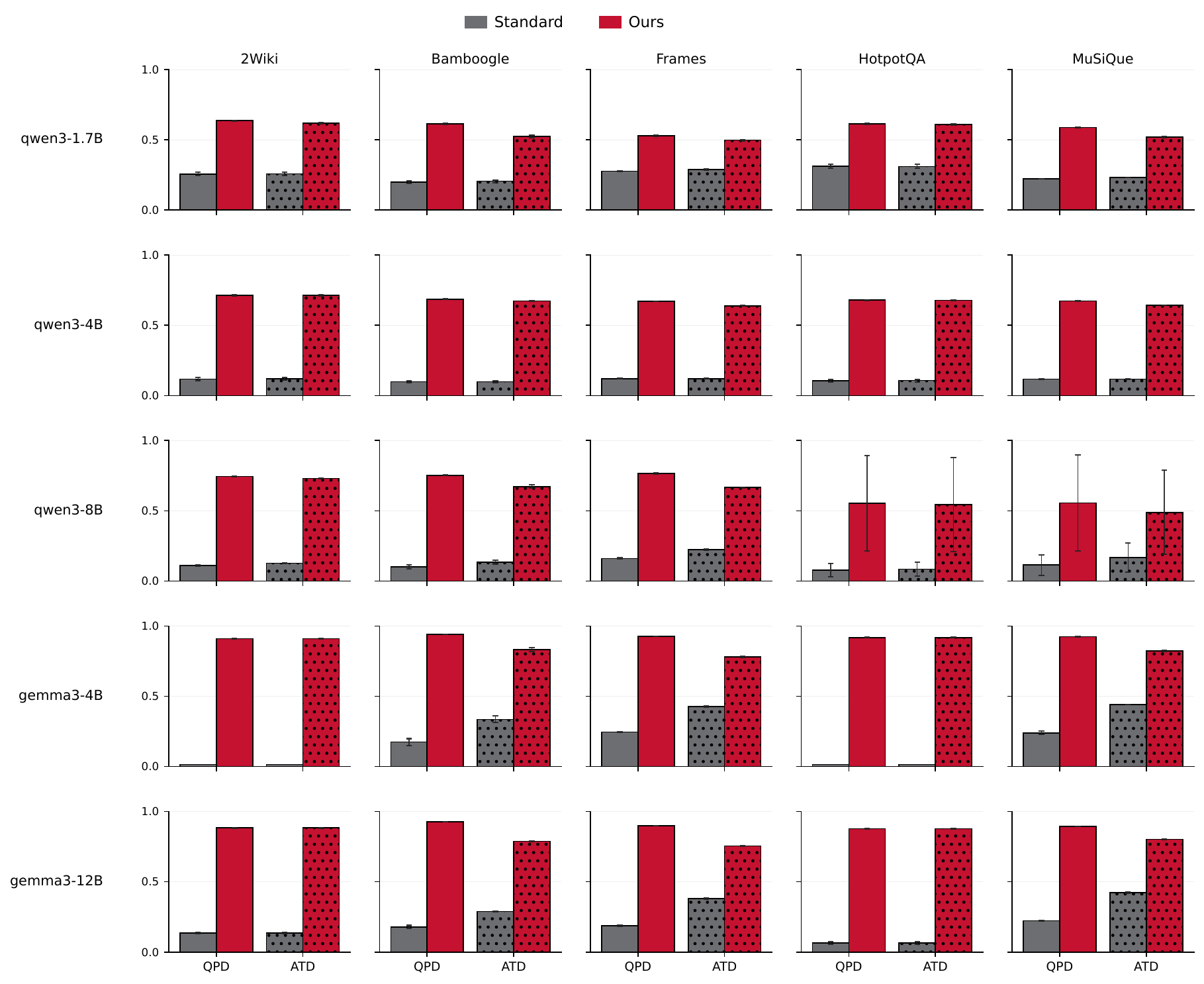}
\caption{Multi-hop QA results, reporting per-cell QPD and ATD comparing standard (gray) and
\textbf{\texttt{DivInit}} (red) at $k{=}4$, across five models and five datasets.}
\label{fig:appfig-mhqa-diag}
\end{figure*}

\begin{figure*}[p]
\centering
\includegraphics[width=0.70\textwidth]{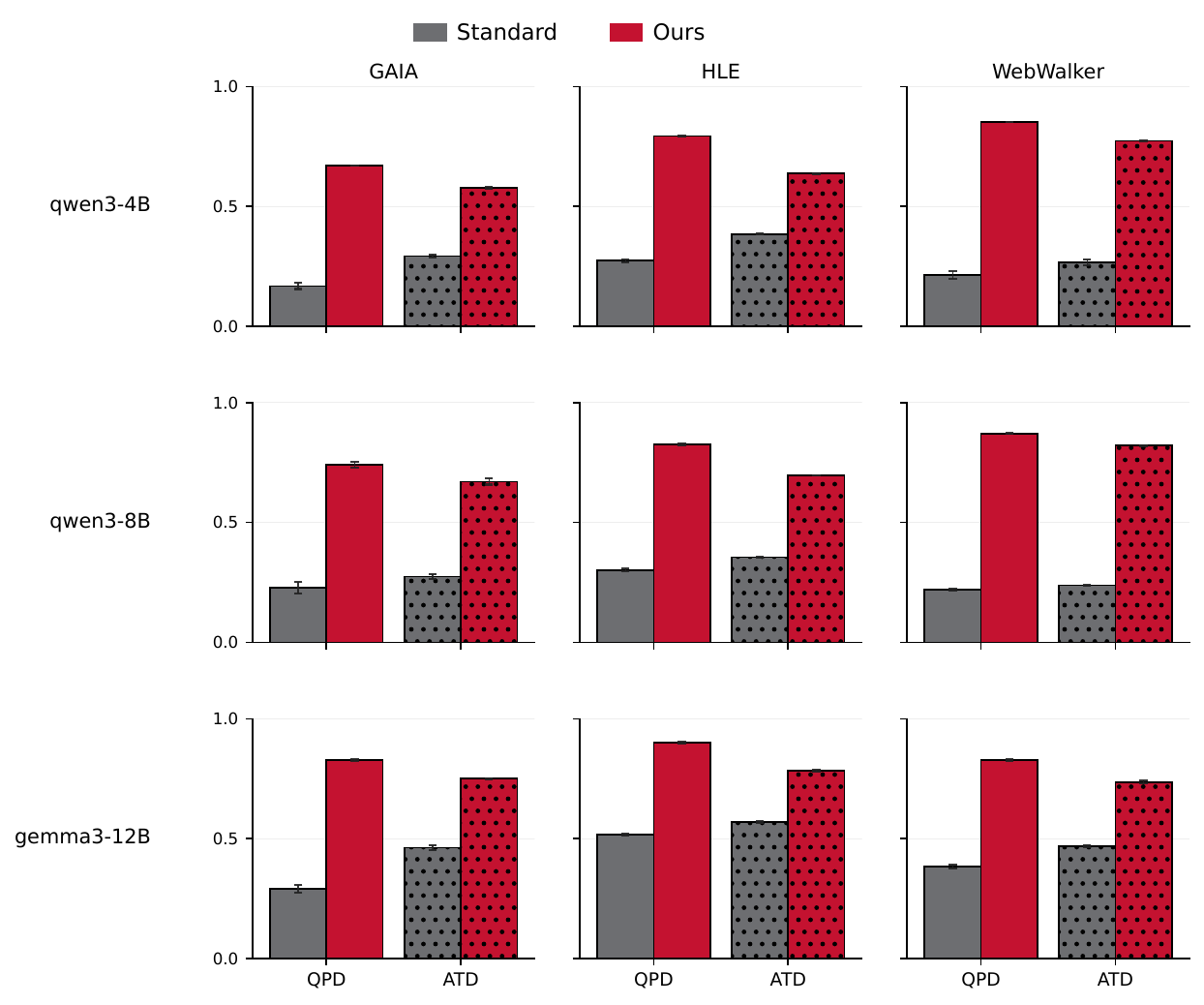}
\caption{Open-web reasoning results, reporting per-cell QPD and ATD comparing standard and \textbf{\texttt{DivInit}}
at $k{=}4$, across three models and three datasets.}
\label{fig:appfig-reasoning-diag}
\end{figure*}

Figures~\ref{fig:appfig-mhqa-passk} to~\ref{fig:appfig-reasoning-diag} report QPD and ATD
alongside pass@$k$ curves over $k \in \{1, 4, 8\}$ for every (model, dataset) cell in both
benchmark groups. The pattern from \S\ref{sec:anchor} is consistent across the grid: standard
parallel sampling concentrates at low QPD, \textbf{\texttt{DivInit}} at high QPD, and ATD
tracks QPD throughout.

\subsection{Aggregation Accuracy}
\label{app:aggregation}

Aggregating $k$ parallel rollouts into a single answer is not a trivial task, with current
strategies falling short of the pass@$k$ ceiling~\citep{li2026benchmark}. Our work focuses on
pass@$k$, leaving aggregation to existing methods. To verify that \textbf{\texttt{DivInit}}'s
gains transfer past the oracle regime, we run AggAgent~\citep{lee2026agentic}, a
state-of-the-art LLM-based aggregator, on both standard and \textbf{\texttt{DivInit}}
trajectories. Table~\ref{tab:agg} reports the results. When compared to standard parallel
sampling \textbf{\texttt{DivInit}} lifts both the pass@4, and AggAgent's pass@1 metrics. The diversity gain carries through to single-answer accuracy, while a gap to the pass@4 ceiling remains under both conditions.

\subsection{First Turn Wall Clock Time}
\label{app:wall_time}

\begin{table}[t]
\centering
\small
\setlength{\tabcolsep}{5pt}
\begin{tabular}{l rr rr rr}
\toprule
 & \multicolumn{2}{c}{Qwen3-1.7B} & \multicolumn{2}{c}{Qwen3-4B} & \multicolumn{2}{c}{Qwen3-8B} \\
\cmidrule(lr){2-3}\cmidrule(lr){4-5}\cmidrule(lr){6-7}
Method & Wall & qps & Wall & qps & Wall & qps \\
\midrule
Standard  & \textbf{96.6}  & \textbf{2.65} & 204.6          & 1.25          & 285.3          & 0.90 \\
DivInit   & 102.4          & 2.50          & \textbf{119.1} & \textbf{2.15} & \textbf{164.9} & \textbf{1.55} \\
\bottomrule
\end{tabular}
\caption{Wall-clock time (seconds) and question throughput for first-turn search-query generation on 256 multi-hop questions, on a single L40S GPU.}
\label{tab:turn1-throughput}
\vspace{-0.25cm}
\end{table}

We batch 256 questions in a single vLLM call and time the first turn under each method. Turns 2 through 8 issue one search per thread under both methods and have identical per-turn cost, so we benchmark turn 1 only.

Table~\ref{tab:turn1-throughput} shows DivInit is slower at the 1.7B parameter scale, and faster at 4B and 8B. Standard runs $k{=}4$ reasoning traces per question, each followed by a single query. In contrast, \textbf{\texttt{DivInit}} runs a single reasoning trace for $n{=}16$ queries. At 4B and 8B, the $k$-fold multiplier on the long reasoning trace dominates, and wall time tracks total tokens. Turns 2 through 8 dilute this turn-1 gap across the full eight-turn rollout.

\section{Case Studies}
\label{sec:appendix-cases}

\newcounter{case}
\renewcommand{\thecase}{C.\arabic{case}}
\newcommand{\caselabel}[1]{\refstepcounter{case}\label{#1}}

We present four trajectory-level case studies from GAIA experiments
with Qwen3-8B at $k{=}4$ and $n{=}16$. Cases~\ref{case:nature2020}
and~\ref{case:nasa} show where \textbf{\texttt{DivInit}} succeeds
while standard parallel sampling fails; Cases~\ref{case:doctorwho}
and~\ref{case:rooster} show the reverse.
\clearpage
\clearpage

\begin{figure*}[p]
\caselabel{case:nature2020}
\small
\begin{tcolorbox}[
  colback=white, colframe=black!40,
  boxrule=0.6pt, arc=3pt,
  left=8pt, right=8pt, top=6pt, bottom=6pt,
  title={\normalsize\textbf{C.1 \ Statistical Significance in Nature 2020}},
  fonttitle=\bfseries, coltitle=black,
  colbacktitle=gray!10,
]

\begin{casebox}
\textbf{Question.}~If we assume all articles published by Nature in
2020 relied on statistical significance to justify their findings and
they on average came to a $p$-value of 0.04, how many papers would be
incorrect as to their claims of statistical significance? Round up to
the next integer.\\[3pt]
\textbf{Gold:}~\texttt{41}\quad
\textbf{Standard QPD:}~0.120\quad
\textbf{DI QPD:}~0.901\quad
\textbf{Standard:}~\xmark\quad
\textbf{DI:}~\cmark
\end{casebox}

\begin{diagbox}
\textbf{Diagnosis.}~A two-step problem: look up the Nature 2020
article count, then apply the false-positive rate. All four standard
threads issue near-identical queries and either loop or mis-apply the
calculation. DivInit seeds probe orthogonal sub-problems, letting one
thread assemble both steps correctly.
\end{diagbox}

\begin{poolbox}
\textbf{Oversampled pool ($n{=}16$; selected in \textbf{bold}):}
\textbf{1.}~Nature total articles 2020 excl.\ book reviews;
2.~common $p$-value thresholds;
3.~understanding $p$-values;
4.~Nature articles by year;
\textbf{5.}~percent findings incorrect at $p{=}0.04$;
6.~2020 Nature by field;
7.~type I error;
8.~$p$-values in claims 2020;
9.~concerns about $p$-values;
10.~Nature 2020 by category;
\textbf{11.}~examples of incorrect significance;
12.~reliable significance methods;
13.~Nature on $p$-value interpretation;
14.~retracted articles for statistical misinterpretation;
15.~$p{=}0.04$ implications;
\textbf{16.}~Nature's policy on significance 2020.
\end{poolbox}

\medskip
\begin{minipage}[t]{0.48\linewidth}
\colhdr{Standard Parallel}{0.120}{Standardtint}

\begin{thr}{Thread 0}
\sq{total number of articles published by Nature in 2020}
\sr{Volumes $\mid$ Nature \dots fewer than 8\% of submitted papers \dots}
\note{T2--T5: Loops on same query; never extracts count.}
\ans{[malformed]}
\end{thr}

\begin{thr}{Thread 1}
\sq{total number of articles published by Nature in 2020}
\sr{Nature publishes about 15 articles per issue \dots}
\note{T2--T3: Finds ``12{,}000'' from aggregator.}
\ans{12000}~\xmark
\end{thr}

\begin{thr}{Thread 2}
\sq{number of articles published by Nature in 2020}
\sr{Article Type Article (1002), Year 2020 (1002) \dots}
\note{Mis-applies false-positive rate as 100\%.}
\ans{1002}~\xmark
\end{thr}

\begin{thr}{Thread 3}
\sq{total number of articles published by Nature in 2020}
\sr{Article (1002), Year 2020 \dots}
\note{Applies 5\% rate without re-reading question.}
\ans{50}~\xmark
\end{thr}

\smallskip
\oraclexmark{All threads fail.}
\end{minipage}%
\hfill%
\begin{minipage}[t]{0.48\linewidth}
\colhdr{\texttt{DivInit}}{0.901}{s3tint}

\begin{thr}{Thread 0 -- article-count seed}
\sq{Nature journal total articles 2020 excl.\ book reviews}
\sr{Nature publishes peer-reviewed research \dots}
\note{T2--T5: Converges on 1002 but cannot finish calculation.}
\ans{[malformed]}
\end{thr}

\begin{thr}{Thread 1 -- false-positive-rate seed}
\sq{percentage of findings incorrect at $p$-value 0.04}
\sr{$p$-value of 0.04 represents a 4\% risk \dots}
\note{T2--T4: Retrieves rate but cannot find article count.}
\ans{[malformed]}
\end{thr}

\begin{thr}{Thread 2 -- policy seed}
\sq{Nature's policy on statistical significance in 2020 articles}
\sr{Scientists rise up against statistical significance \dots}
\note{T2--T5: Pivots to article count but loops.}
\ans{[malformed]}
\end{thr}

\begin{thr}{Thread 3 -- incorrect-significance seed}
\sq{examples of incorrect statistical significance in published research}
\sr{Common statistical errors \dots}
\sq{number of articles published by Nature in 2020}
\sr{Article (1002) \dots $p{=}0.04 \Rightarrow 4\%$ false-positive rate \dots}
\note{Combines: $1002\times0.04=40.08\Rightarrow\lceil40.08\rceil=41$.}
\answin{41}~\cmark
\end{thr}

\smallskip
\oraclecmark{Thread 3 assembles both steps correctly.}
\end{minipage}

\medskip
\begin{diagbox}
\textbf{Mechanism.}~Standard seeds are lexical variants of the same
article-count query; none frames the false-positive sub-problem.
DivInit seeds target orthogonal aspects, giving Thread~3 the framing
to combine both retrieved facts.
\end{diagbox}

\end{tcolorbox}
\end{figure*}

\begin{figure*}[p]
\caselabel{case:nasa}
\small
\begin{tcolorbox}[
  colback=white, colframe=black!40,
  boxrule=0.6pt, arc=3pt,
  left=8pt, right=8pt, top=6pt, bottom=6pt,
  title={\normalsize\textbf{C.2 \ NASA Award Number Chain}},
  fonttitle=\bfseries, coltitle=black,
  colbacktitle=gray!10,
]

\begin{casebox}
\textbf{Question.}~On June 6, 2023, an article by Carolyn Collins
Petersen was published in Universe Today. This article mentions a team
that produced a paper about their observations, linked at the bottom of
the article. Find this paper. Under what NASA award number was the work
performed by R.\,G.~Arendt supported?\\[3pt]
\textbf{Gold:}~\texttt{80GSFC21M0002}\quad
\textbf{Standard QPD:}~0.098\quad
\textbf{DI QPD:}~0.743\quad
\textbf{Standard:}~\xmark\quad
\textbf{DI:}~\cmark
\end{casebox}

\begin{diagbox}
\textbf{Diagnosis.}~A multi-hop chain: locate the Universe Today
article, follow the linked paper, then find the funding record for a
specific author. All four standard threads pack every keyword into one
query and retrieve the same wrong award number from a different paper.
DivInit decompose the chain; three threads independently reach
the correct record.
\end{diagbox}

\begin{poolbox}
\textbf{Selected seeds (from $n{=}16$):}
\textbf{(1)}~Funding source for R.\,G.\ Arendt research Universe Today June 6, 2023;
\textbf{(2)}~Paper linked in Universe Today article Carolyn Collins Petersen June 6, 2023;
\textbf{(3)}~NASA grant number R.\,G.\ Arendt work cited Universe Today June 2023;
\textbf{(4)}~NASA award number research cited Universe Today article June 6, 2023;
5.~Carolyn Collins Petersen Universe Today astrophysics June 2023;
6.~R.\,G.\ Arendt NASA funding acknowledgment 2023;
7.~Universe Today June 2023 linked paper authors;
8.~galactic center observations paper June 2023 NASA grant.
\end{poolbox}

\medskip
\begin{minipage}[t]{0.48\linewidth}
\colhdr{Standard Parallel}{0.098}{Standardtint}

\begin{thr}{Thread 0}
\sq{Carolyn Collins Petersen June 6 2023 Universe Today R.\,G.\ Arendt NASA award}
\sr{NASA award NNX15AF61G \dots}
\note{Retrieves wrong award from a different paper.}
\ans{NNX15AF61G}~\xmark
\end{thr}

\begin{thr}{Thread 1}
\sq{Carolyn Collins Petersen June 6 2023 Universe Today R.\,G.\ Arendt NASA award number}
\sr{NASA award NNX15AF61G \dots}
\note{Same result; budget exhausted.}
\ans{NNX15AF61G}~\xmark
\end{thr}

\begin{thr}{Thread 2}
\sq{Carolyn Collins Petersen June 6 2023 Universe Today R.\,G.\ Arendt NASA award number}
\sr{NASA award NNX15AF61G \dots}
\note{Identical retrieval path.}
\ans{NNX15AF61G}~\xmark
\end{thr}

\begin{thr}{Thread 3}
\sq{Carolyn Collins Petersen June 6 2023 Universe Today R.\,G.\ Arendt NASA award number}
\sr{NASA award NNX15AF61G \dots}
\note{Identical retrieval path.}
\ans{NNX15AF61G}~\xmark
\end{thr}

\smallskip
\oraclexmark{All threads return the wrong award number.}
\end{minipage}%
\hfill%
\begin{minipage}[t]{0.48\linewidth}
\colhdr{\texttt{DivInit}}{0.743}{s3tint}

\begin{thr}{Thread 0 -- funding seed}
\sq{Funding source for R.\,G.\ Arendt research in Universe Today June 6, 2023}
\sr{NASA award 80GSFC21M0002 \dots}
\answin{80GSFC21M0002}~\cmark
\end{thr}

\begin{thr}{Thread 1 -- article-first seed}
\sq{Paper linked in Universe Today article Carolyn Collins Petersen June 6, 2023}
\sr{Paper on galactic center observations \dots}
\note{T2--T3: Locates paper but wrong funding entry.}
\ans{80NSSC22K0568}~\xmark
\end{thr}

\begin{thr}{Thread 2 -- grant seed}
\sq{NASA grant number R.\,G.\ Arendt work cited Universe Today June 2023}
\sr{NASA award 80GSFC21M0002 \dots}
\answin{80GSFC21M0002}~\cmark
\end{thr}

\begin{thr}{Thread 3 -- award seed}
\sq{NASA award number research cited Universe Today article June 6, 2023}
\sr{NASA award 80GSFC21M0002 \dots}
\answin{80GSFC21M0002}~\cmark
\end{thr}

\smallskip
\oraclecmark{Three threads independently reach the correct award.}
\end{minipage}

\medskip
\begin{diagbox}
\textbf{Mechanism.}~Standard threads overload the query with every
entity and land on a prominent but wrong record. DivInit seeds approach
the chain from different entry points; three of four phrase the funding
lookup in a way that surfaces the correct paper and award.
\end{diagbox}

\end{tcolorbox}
\end{figure*}

\begin{figure*}[p]
\caselabel{case:doctorwho}
\small
\begin{tcolorbox}[
  colback=white, colframe=black!40,
  boxrule=0.6pt, arc=3pt,
  left=8pt, right=8pt, top=6pt, bottom=6pt,
  title={\normalsize\textbf{C.3 \ Doctor Who Script Heading}},
  fonttitle=\bfseries, coltitle=black,
  colbacktitle=gray!10,
]

\begin{casebox}
\textbf{Question.}~In Series 9, Episode 11 of Doctor Who, the Doctor
is trapped inside an ever-shifting maze. What is this location called
in the official script for the episode? Give the setting exactly as it
appears in the first scene heading.\\[3pt]
\textbf{Gold:}~\texttt{THE CASTLE}\quad
\textbf{Standard QPD:}~0.183\quad
\textbf{DI QPD:}~0.631\quad
\textbf{Standard:}~\cmark\quad
\textbf{DI:}~\xmark
\end{casebox}

\begin{diagbox}
\textbf{Diagnosis.}~A direct script lookup. Standard Thread~1 happens
to retrieve the correct answer through a broad query. DivInit seeds
push toward conceptual sub-problems; all four threads land on
plot-summary sources that return the episode title instead of the
in-script heading.
\end{diagbox}

\begin{poolbox}
\textbf{Selected seeds (from $n{=}16$):}
\textbf{(1)}~Doctor Who S9E11 official script first scene heading location;
\textbf{(2)}~exact setting name Doctor Who Series 9 Episode 11 maze script;
\textbf{(3)}~Doctor Who Heaven Sent script PDF scene heading;
\textbf{(4)}~Doctor Who S9E11 script download castle location name;
5.~Doctor Who Series 9 Episode 11 maze setting name;
6.~Heaven Sent Doctor Who location inside maze official name;
7.~Doctor Who S9E11 castle setting script heading text;
8.~BBC script Doctor Who Heaven Sent first scene.
\end{poolbox}

\medskip
\begin{minipage}[t]{0.48\linewidth}
\colhdr{Standard Parallel}{0.183}{Standardtint}

\begin{thr}{Thread 0}
\sq{Doctor Who Series 9 Episode 11 location name}
\note{Retrieves episode title from plot summaries.}
\ans{Heaven Sent}~\xmark
\end{thr}

\begin{thr}{Thread 1}
\sq{unknown language article DDC 633 Bielefeld University BASE 2020}
\sr{\dots exact setting name in script \dots THE CASTLE \dots}
\answin{THE CASTLE}~\cmark
\end{thr}

\begin{thr}{Thread 2}
\sq{Doctor Who Series 9 Episode 11 first scene heading}
\note{Finds fandom wikis; returns episode title.}
\ans{Heaven Sent}~\xmark
\end{thr}

\begin{thr}{Thread 3}
\sq{Doctor Who Series 9 Episode 11 maze location}
\note{Identical retrieval path to Thread 0.}
\ans{Heaven Sent}~\xmark
\end{thr}

\smallskip
\oraclecmark{Thread 1 incidentally retrieves the correct heading.}
\end{minipage}%
\hfill%
\begin{minipage}[t]{0.48\linewidth}
\colhdr{\texttt{DivInit}}{0.631}{s3tint}

\begin{thr}{Thread 0 -- scene-heading seed}
\sq{Doctor Who S9E11 official script first scene heading location}
\sr{Script PDF \dots INT.\ THE CASTLE \dots}
\note{Locates script but returns episode title on answer turn.}
\ans{Heaven Sent}~\xmark
\end{thr}

\begin{thr}{Thread 1 -- exact-setting seed}
\sq{exact setting name Doctor Who Series 9 Episode 11 maze script}
\sr{Fandom wiki \dots Heaven Sent episode \dots}
\ans{Heaven Sent}~\xmark
\end{thr}

\begin{thr}{Thread 2 -- script PDF seed}
\sq{Doctor Who Heaven Sent script PDF scene heading}
\sr{Script \dots first heading: THE CASTLE \dots}
\note{Retrieves script but mis-formats answer.}
\ans{The Castle}~\xmark
\end{thr}

\begin{thr}{Thread 3 -- download seed}
\sq{Doctor Who S9E11 script download castle location name}
\sr{Heaven Sent script \dots INT.\ THE CASTLE \dots}
\note{Finds script; answers with episode title instead.}
\ans{Heaven Sent}~\xmark
\end{thr}

\smallskip
\oraclexmark{All DivInit threads miss the exact capitalised heading.}
\end{minipage}

\medskip
\begin{diagbox}
\textbf{Mechanism.}~Diverse seeds collectively land on sources that
describe the episode rather than reproduce the script heading verbatim.
Standard Thread~1's unstructured query happens to surface a page
containing the exact text. This is a failure mode in which query
diversity does not help when the answer requires an exact string match
from a specific document.
\end{diagbox}

\end{tcolorbox}
\end{figure*}

\begin{figure*}[p]
\caselabel{case:rooster}
\small
\begin{tcolorbox}[
  colback=white, colframe=black!40,
  boxrule=0.6pt, arc=3pt,
  left=8pt, right=8pt, top=6pt, bottom=6pt,
  title={\normalsize\textbf{C.4 \ Rooster and Hamster Song Composer}},
  fonttitle=\bfseries, coltitle=black,
  colbacktitle=gray!10,
]

\begin{casebox}
\textbf{Question.}~Who composed the song that was performed by a
rooster and a hamster in separate animated videos at separate tempos
with different lyrics? Answer using the format ``First Last''.\\[3pt]
\textbf{Gold:}~\texttt{Roger Miller}\quad
\textbf{Standard QPD:}~0.241\quad
\textbf{DI QPD:}~0.784\quad
\textbf{Standard:}~\xmark\quad
\textbf{DI:}~\cmark
\end{casebox}

\begin{diagbox}
\textbf{Diagnosis.}~Standard threads retrieve results about Hampton
the Hamster, a viral internet character, and never connect the rooster
video to identify the original song. Two DivInit threads approach the
problem from different angles and independently identify Roger Miller
as the composer.
\end{diagbox}

\begin{poolbox}
\textbf{Selected seeds (from $n{=}16$):}
\textbf{(1)}~animated rooster video song original composer viral;
\textbf{(2)}~hamster rooster same song different tempo animated videos composer;
\textbf{(3)}~Roger Miller Robin Hood song rooster hamster animated cover;
\textbf{(4)}~original song covered by animated rooster and hamster internet videos;
5.~Hampton the Hamster original song composer;
6.~rooster animated music video viral internet song;
7.~Robin Hood whistle-stop song animated animals;
8.~viral hamster song same melody rooster video origin.
\end{poolbox}

\medskip
\begin{minipage}[t]{0.48\linewidth}
\colhdr{Standard Parallel}{0.241}{Standardtint}

\begin{thr}{Thread 0}
\note{No query issued on turn 1; malformed output.}
\ans{[malformed]}
\end{thr}

\begin{thr}{Thread 1}
\sq{song performed by rooster and hamster animated videos separate tempos}
\sr{Hampton the Hamster \dots viral character \dots}
\ans{Hampton}~\xmark
\end{thr}

\begin{thr}{Thread 2}
\sq{song performed by rooster and hamster animated videos different tempos}
\sr{Hampton The Hamster \dots}
\ans{Hampton The Hamster}~\xmark
\end{thr}

\begin{thr}{Thread 3}
\sq{song performed by rooster and hamster animated videos}
\sr{Hampton The Hamster \dots Ray Stevens \dots}
\ans{Ray Stevens}~\xmark
\end{thr}

\smallskip
\oraclexmark{All threads fixate on the hamster character; none identifies the composer.}
\end{minipage}%
\hfill%
\begin{minipage}[t]{0.48\linewidth}
\colhdr{\texttt{DivInit}}{0.784}{s3tint}

\begin{thr}{Thread 0 -- rooster-first seed}
\sq{animated rooster video song original composer viral}
\sr{Robin Hood rooster animation \dots Roger Miller \dots}
\answin{Roger Miller}~\cmark
\end{thr}

\begin{thr}{Thread 1 -- connection seed}
\sq{hamster rooster same song different tempo animated videos composer}
\sr{Hampton Hamster \dots Michael Stipe \dots}
\ans{Michael Stipe}~\xmark
\end{thr}

\begin{thr}{Thread 2 -- composer seed}
\sq{Roger Miller Robin Hood song rooster hamster animated cover}
\sr{Roger Miller composer \dots Robin Hood whistle-stop \dots}
\answin{Roger Miller}~\cmark
\end{thr}

\begin{thr}{Thread 3 -- original-song seed}
\sq{original song covered by animated rooster and hamster internet videos}
\sr{Walter Scharf \dots Robin Hood \dots}
\ans{Walter Scharf}~\xmark
\end{thr}

\smallskip
\oraclecmark{Threads 0 and 2 independently identify Roger Miller.}
\end{minipage}

\medskip
\begin{diagbox}
\textbf{Mechanism.}~Standard threads all anchor on the hamster
character and never reach the rooster side of the connection. DivInit
seeds approach the question from different directions; two threads
phrase queries in ways that surface the original song and its composer.
\end{diagbox}

\end{tcolorbox}
\end{figure*}
\end{document}